\documentclass[titlepage]{article} 

\usepackage[utf8]{inputenc} 

\usepackage{geometry} 
\usepackage{graphicx} 
\usepackage{caption}
\usepackage{subcaption}

\usepackage{booktabs} 
\usepackage{array} 
\usepackage{paralist} 
\usepackage{verbatim} 

\usepackage{fancyhdr} 
\usepackage{sectsty}
\allsectionsfont{\sffamily\mdseries\upshape} 

\usepackage[nottoc,notlof,notlot]{tocbibind} 
\usepackage[titles,subfigure]{tocloft} 

\title{Interpreting Deep Neural Networks with Nonlinear Saliency Maps}
\author{Jan Rosenzweig$^{1 \dagger}$, Zoran Cvetkovi\'c$^{1}$, Ivana Rosenzweig$^{2}$\\ 1 Department of Engineering, King's College London, Strand, London
WC2R 2LS\\ 2 Department of Neuroimaging, King's College London, De Crespigny Park, SE5 8AF\\$\dagger$ Corresponding author \\ \\ \\ \\ \\ \\ \\ \\ Corresponding author:\\ Jan Rosenzweig, \\ Department of Engineering, King's College London, Strand, London\\e-mail: jan.1.rosenzweig@kcl.ac.uk }
\date{} 

\begin{document}
\maketitle

\begin{abstract}

A fundamental bottleneck in utilising complex machine learning systems for critical applications has been not knowing why they do and what they do, thus preventing the development of any crucial safety protocols. To date, no method exist that can provide full insight into the granularity of the neural network’s decision process. 
In past, saliency maps were an early attempt at resolving this problem through sensitivity calculations, whereby dimensions of a data point are selected based on how sensitive the output of the system is to them. However, the success of saliency maps has been at best limited, mainly due to the fact that they interpret the underlying learning system through a linear approximation.
We present a novel class of methods for generating nonlinear saliency maps which fully account for the nonlinearity of the underlying learning system. While agreeing with linear saliency maps on simple problems where linear saliency maps are correct, they clearly identify more specific drivers of classification on complex examples where nonlinearities are more pronounced.
In comparison to GradCAM, SmoothGrad, Occlusion Sensitivity and Vanilla Gradient, our nonlinear gradient method has shown equal or better specificity for explaining the classification, especially so for classes other than the top ranked class by probability.
This new class of methods significantly aids interpretability of deep neural networks and related machine learning systems. Crucially, they provide a starting point for their more broad use in serious applications, where  'why' is equally important as 'what'.
\end{abstract}

\section*{Main}

One of the key barriers limiting the mass adoption of machine learning in serious applications is the lack of interpretability of such methods. In layman's terms, machine learning is quite good at providing the answers along the lines of "computer says no", but it is quite difficult to find out "why not". 

This is especially important when any decisions made by such systems have serious consequences. In a well publicized case, Amazon was forced to abandon their AI recruitment tool when it turned out it had a clear bias against female candidates \cite{amazon}. Similar anecdotes revolving around machine learning systems behaving in unintended ways are a daily staple in the world of AI and machine learning, and most researchers have encountered numerous examples, ranging from amusing to distasteful. Biases in AI and machine learning have even been addressed by the World Economic Forum \cite{wef}.

In essence, deep learning systems are an example of black-box systems, providing answers but no insights into how these answers were reached.

Saliency maps are a decade-old tool for interpreting deep learning systems. In computer vision and related applications, they are used to highlight the areas of the image that are most salient to its classification, which has since been generalized to other, non-vision related deep neural networks (DNN)  \cite{sal1}.

At their core, saliency maps rely on the simple  calculation of gradients. The sensitivity of the classification  (output) to the image (input) is measured by the gradient of the output with respect to the input. Hence traditional saliency maps work by calculating this gradient, and then selecting the regions of the input that  contribute most to the gradient. This is in no small part aided by the fact that deep learning systems already calculate all gradients through backpropagation, and hence the resulting gradients are easily available at zero or minimal additional computational cost. 

This strength of saliency maps as an interpretation tool is also their greatest weakness. A deep neural network implements a nonlinear mapping of the input to the output. The gradient, however, only calculates the linear tangent to this mapping, evaluated at the input. In essence, it calculates the linear component of the DNN as a linear regressor \cite{sal2}. 

In numerous cases, the tangent approximation is sufficient, which has lead to the rapid adoption of saliency maps as an early DNN interpretation tool. 

In  many other  cases, however,  the nonlinearities of the DNN are such that the linear tangent is a poor approximation or, worse, misleading. This is especially noticeable in attempts to use DNNs to classify medical images\cite{med1}; while saliency maps for simple cases indeed highlight the correct region of the image, indicating that they are classified more or less linearly, more complex cases generate saliency maps that invariably pick out wrong parts of the image, indicating that their classification significantly depends on the nonlinearity of the DNN\cite{med2}.

 This puts a fundamental ceiling on the applicability of saliency maps to DNN interpretation.  Consequently, saliency maps have fallen out of favour as a DNN interpretation tool, and their use these days is mostly limited to research purposes \cite{sal2, sal3}.

In this work, we describe a method to overcome these difficulties, by moving away from saliency maps as a linear tangent approximation, towards a fully nonlinear mapping from the input to the output. 

Graphically, a saliency map can be interpreted as deselecting the pixels that do not contribute significantly to the classification, and that can therefore be removed from the image without affecting its classification. In practice, this is usually implemented by masking the non-highlighted pixels with a background colour. The masking is intended to filter out competing classifications, with the effect of increasing the probability of the target classification. 

 This, however, does not have the desired effect. The masking itself introduces new information to the image, and that new information competes with the desired classification. Generally speaking, if more than 20\% of the pixels have been masked, the image is invariably  classified as a "piece of a puzzle" or equivalent, regardless of what is depicted in it.

We modify this approach two-fold, as described in the Methods section. First, we blur the non-selected pixels by subjecting them to a low-pass filter. We then enhance the selected in the direction of the gradient. The gradient is then re-calculated, and the procedure is repeated iteratively.

The motivation for the blurring step lies in the adverse effects of masking, as described above. Filtering is the simplest way to remove information content, without inadvertently introducing additional information, which can happen with masking. It is not the only such step; for instance, random perturbation would have a similar effect, but it would not be repeatable due to randomness.

The motivation for enhancement in the gradient direction comes from the simple gradient descent method; it is the simplest way to select a direction in which the selected class probability increases.

The key difference between our nonlinear saliency maps and linear saliency maps is the updating of the gradient at each step; modifying our method to re-use the same gradient, calculated at the original image, at each step, would revert to a linear saliency map. The difference between the linear and nonlinear saliency map at any given point is therefore directly linked to the local nonlinearity of the DNN at that point.

The specific optimization procedure described in this paper does not converge generally, due to  irreversibility of the low-pass filtering operation. Once the local maximum is overshot by blurring salient pixels, it is impossible for the optimization to un-blur them in order to improve the approximation.  

This lack of convergence does not pose a serious problem. Even though the maximum is never reached, the step parameters can be chosen so that the trajectory passes sufficiently close to  it, before possibly subsequently diverging. Hence the only remaining step is to re-trace the trajectory and  return the points on the trajectory that maximised the target class probability, as various approximations of the relevant maximum. 

We have performed this procedure over a range of well-known DNNs such as VGG16 \cite{vgg}, ResNet50 \cite{resnet}, MobileNet \cite{mobilenet}, InceptionV3 \cite{inception} and Xception \cite{xception}, using mostly images from Imagenet ILSRVC 2017 (\texttt{https://image-net.org}), and others.

\begin{figure}[ht]
     \begin{subfigure}[t]{.5\textwidth}
        \centering
        \includegraphics[width=1\linewidth]{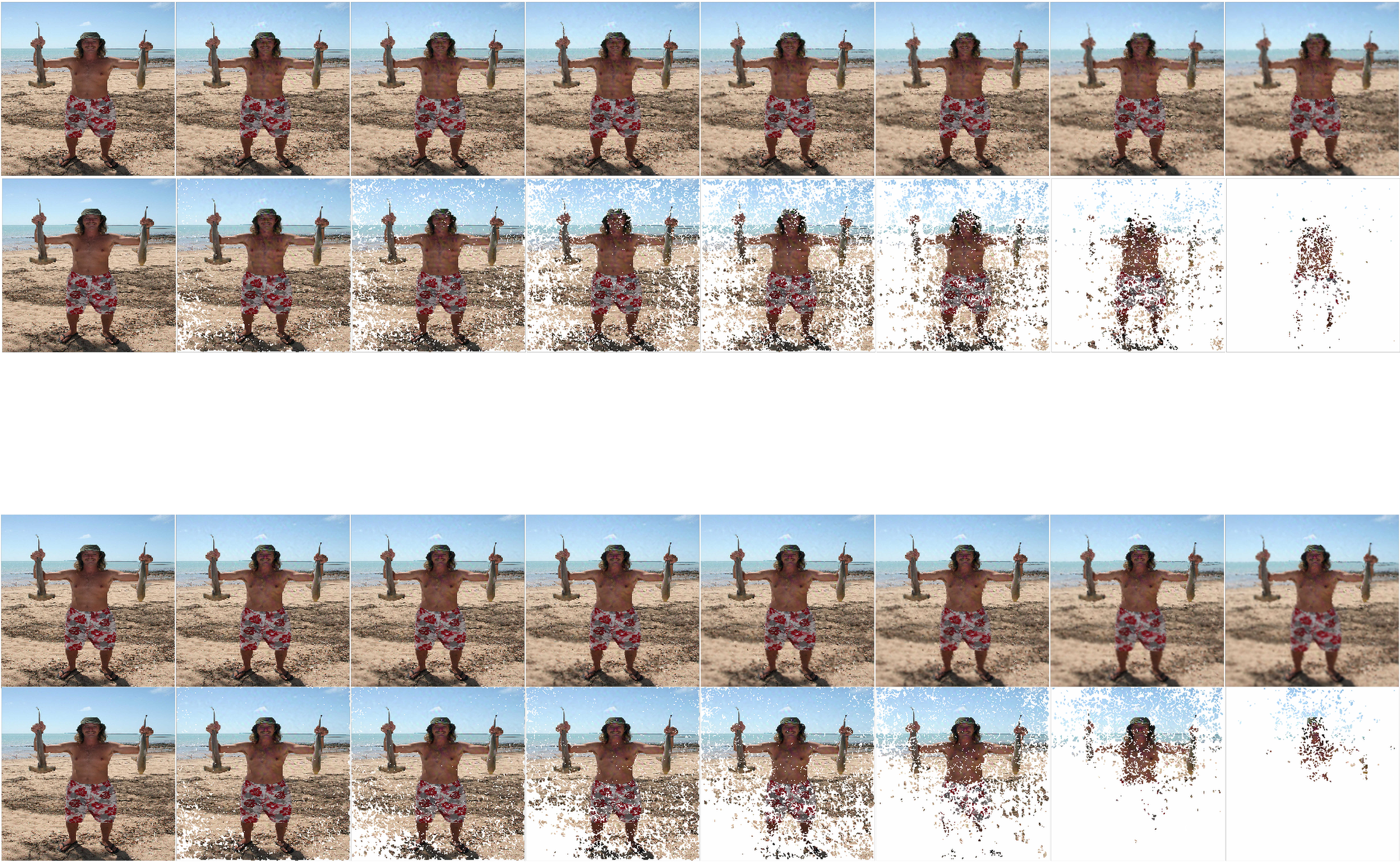}
       \caption{90.10\% \texttt{swimming\_trunks}, 8.03\% \texttt{snorkel}}
    \end{subfigure}
     \begin{subfigure}[t]{.5\textwidth}
        \centering
        \includegraphics[width=1\linewidth]{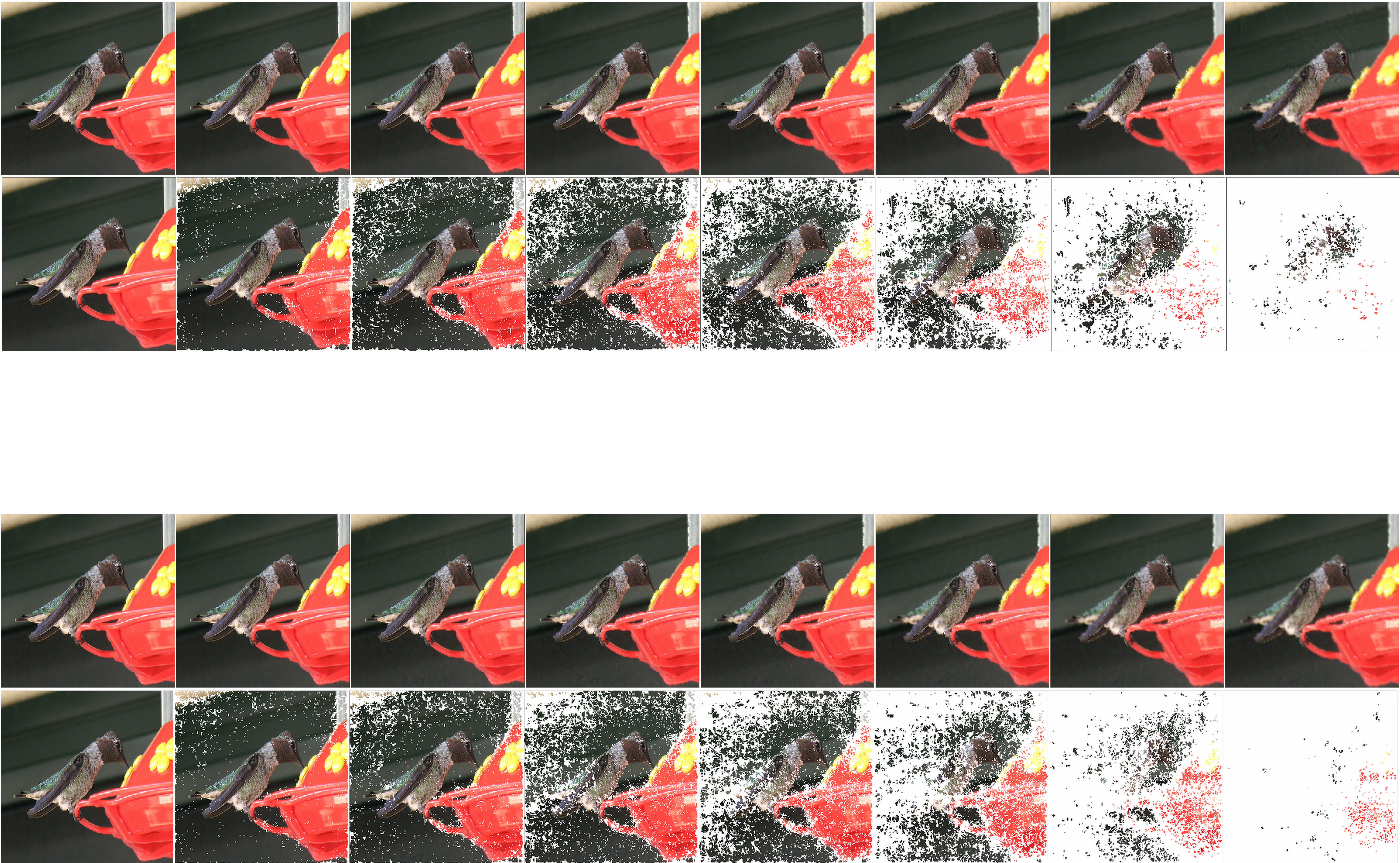}
       \caption{99.85\% \texttt{hummingbird}, 0.06\% \texttt{whistle}}
    \end{subfigure}
     \begin{subfigure}[t]{.5\textwidth}
        \centering
        \includegraphics[width=1\linewidth]{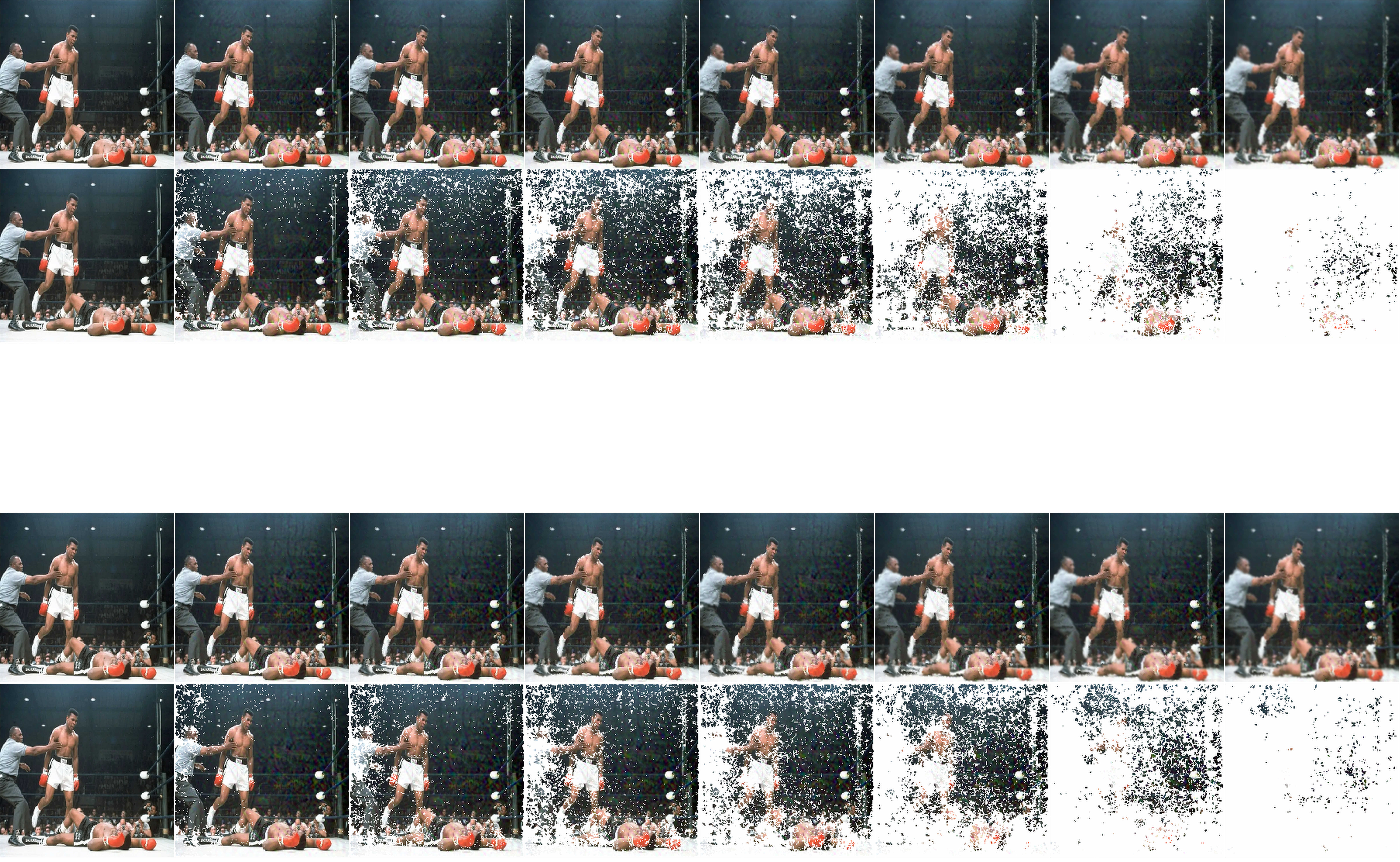}
       \caption{32.85\% \texttt{soccer\_ball}, 18.45\% \texttt{volleyball}}
    \end{subfigure}
     \begin{subfigure}[t]{.5\textwidth}
        \centering
        \includegraphics[width=1\linewidth]{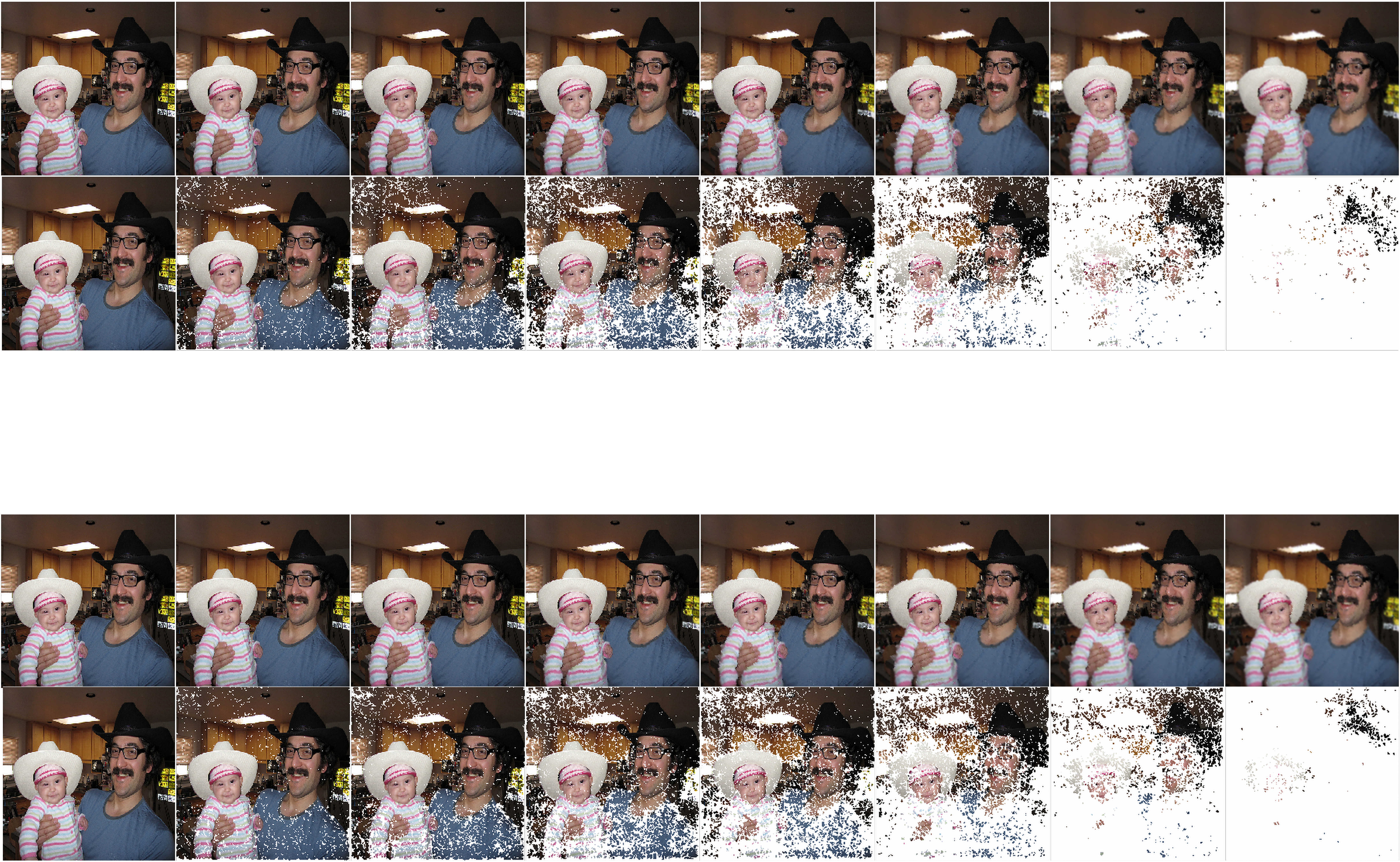}
       \caption{99.99\% \texttt{cowboy\_hat}, 0.01\% \texttt{sombrero}}
    \end{subfigure}
     \begin{subfigure}[t]{.5\textwidth}
        \centering
        \includegraphics[width=1\linewidth]{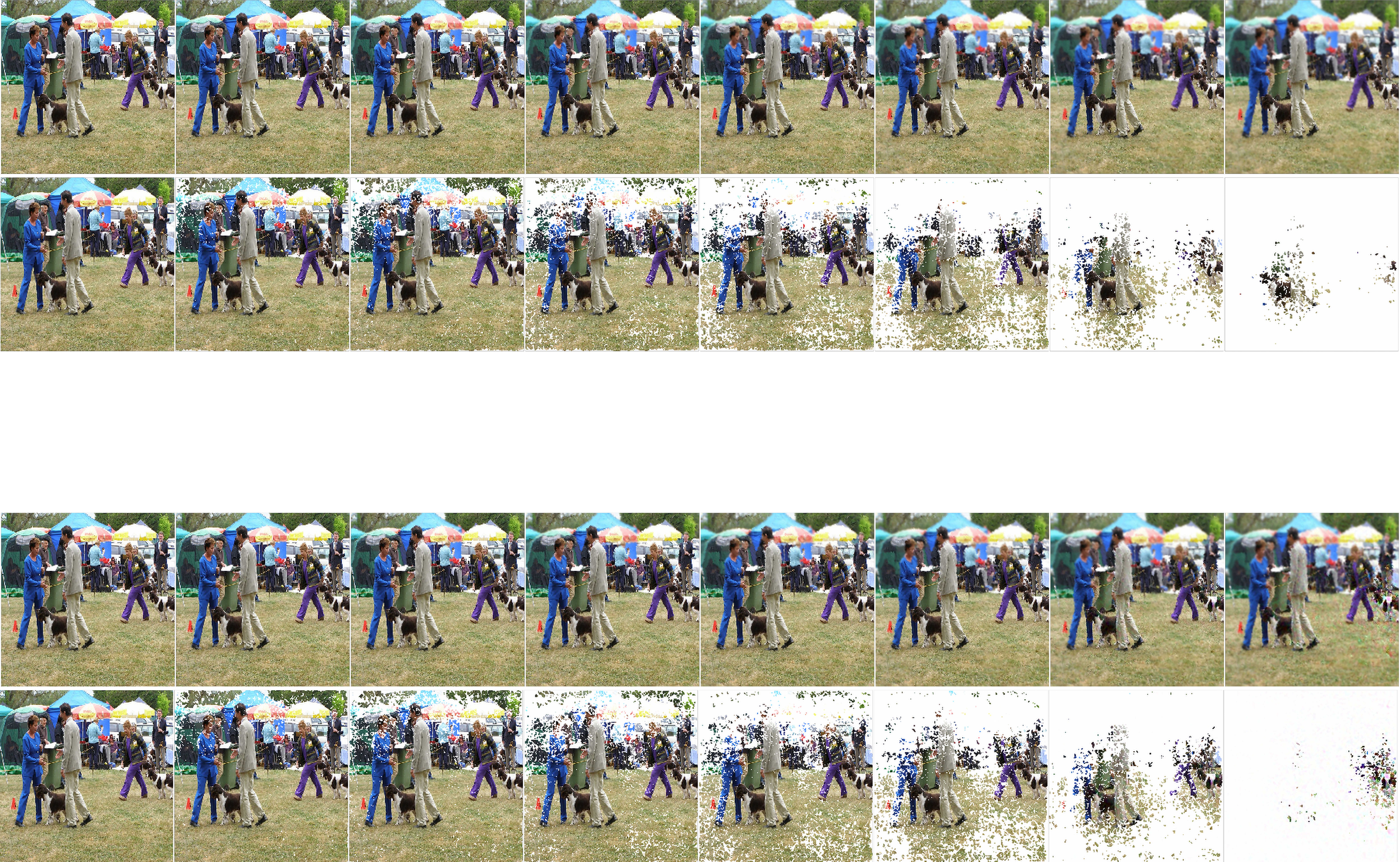}
       \caption{99.98\% \texttt{English\_springer}, 0.01\% \texttt{Border\_collie}}
    \end{subfigure}    
     \begin{subfigure}[t]{.5\textwidth}
        \centering
        \includegraphics[width=1\linewidth]{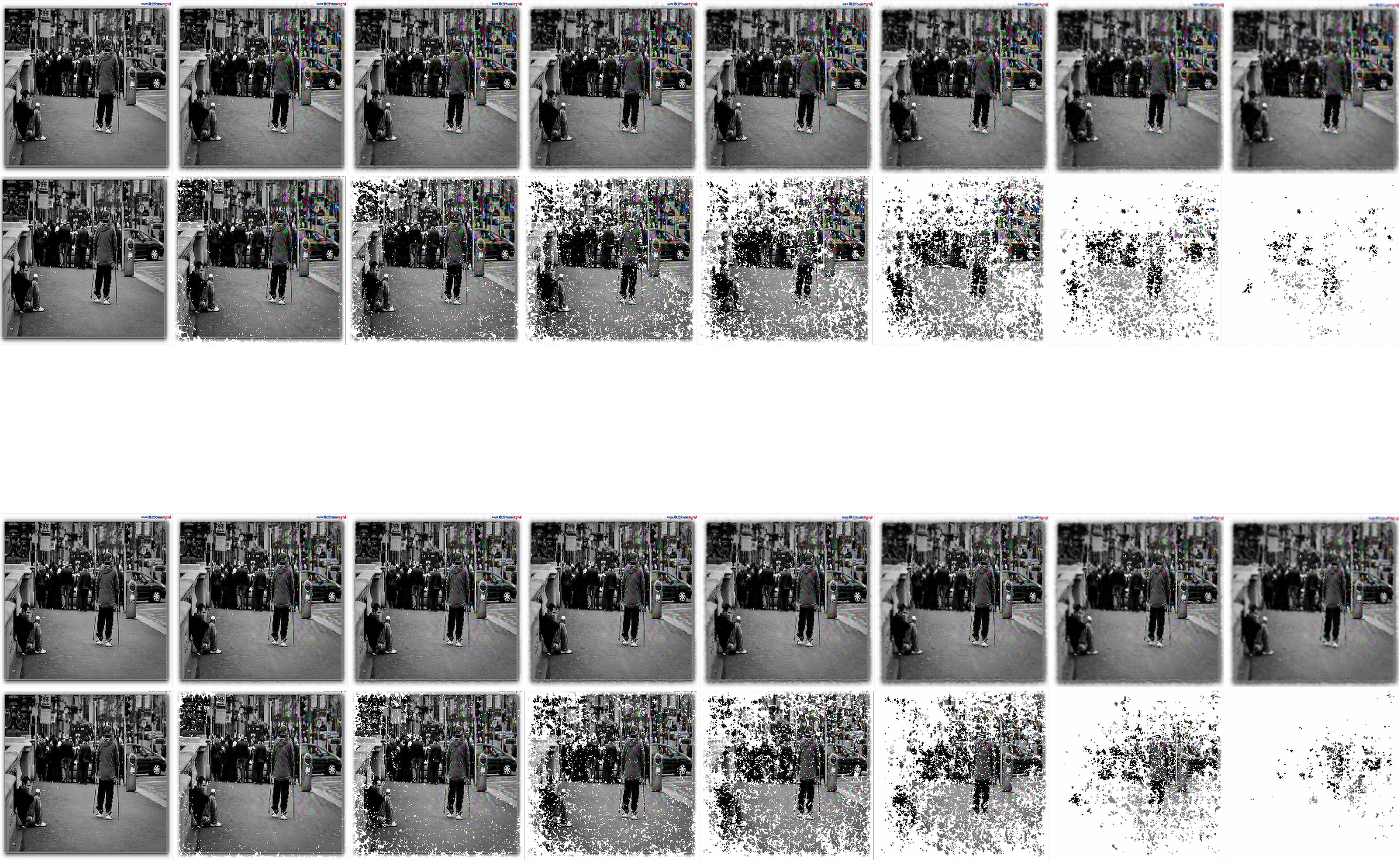}
       \caption{40.89\% \texttt{shoe\_shop}, 22.80\% \texttt{turnstile}}
    \end{subfigure}
     \begin{subfigure}[t]{.5\textwidth}
        \centering
        \includegraphics[width=1\linewidth]{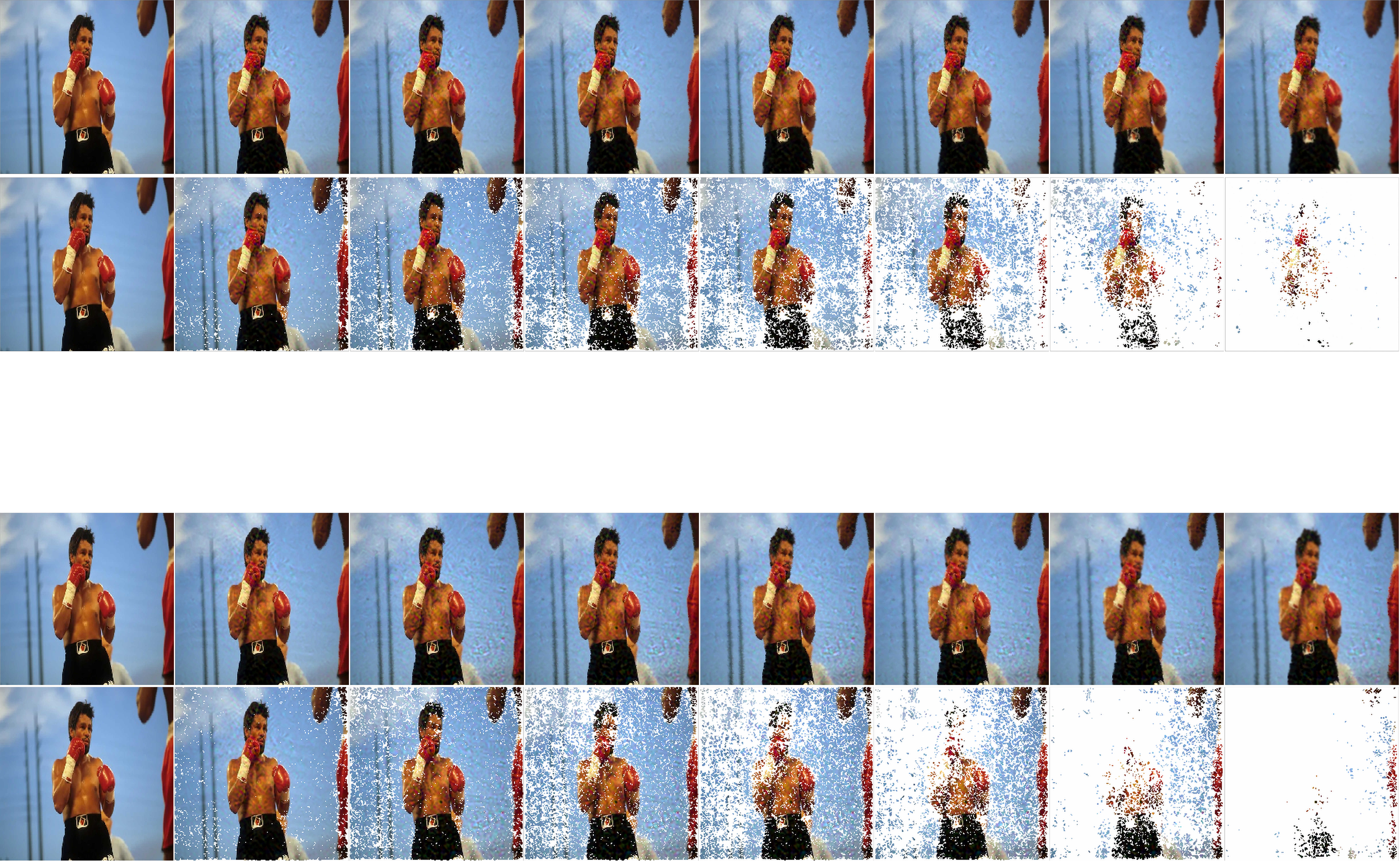}
       \caption{92.29\% \texttt{ice\_lolly},  6.14\% \texttt{swimming\_trunks}}
    \end{subfigure}
     \begin{subfigure}[t]{.5\textwidth}
        \centering
        \includegraphics[width=1\linewidth]{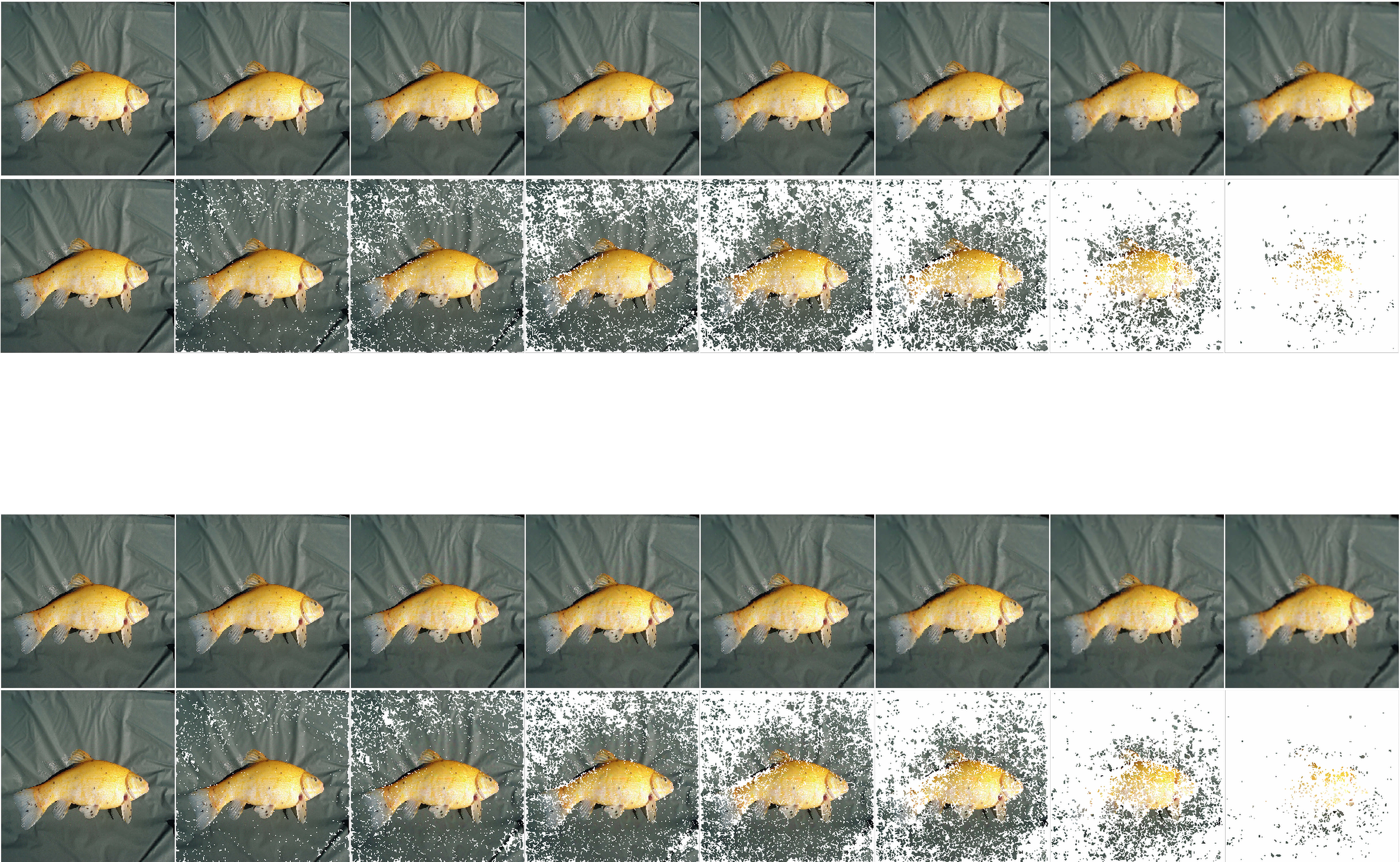}
       \caption{99.70\% \texttt{tench}, 0.27\% \texttt{goldfish}}
    \end{subfigure}
\caption{Image modifications on eight examples, using the Xception DNN for the highest and second highest ranked categories, with masked saliency maps below. The snapshots are, from left to right, with 0\%, 5\%, 10\%, 20\%, 30\%, 50\%, 75\% and 95\% of pixels blurred / masked.}\label{fig:images}
\end{figure}

The results are striking. As  shown in Figure \ref{fig:images}, we can work with  classes that score as low as 0.01\% probability for a given image, and enhance the image to classify in such a class with probability higher than 95\%. We thereby come as close to a pure class representation as is reasonably possible for a given image.  By subsequently masking the non-highlighted pixels, we obtain the equivalent masked saliency map, but this time reflecting the full nonlinearity of the DNN, instead of only its tangent. Trajectories of nonlinear saliency maps are shown alongside of those applying the equivalent transformation according to the linear saliency map (where the gradient is calculated at the original image and never modified), in order to facilitate a direct comparison.

\begin{figure}[ht]
     \begin{subfigure}[t]{.5\textwidth}
        \centering
        \includegraphics[width=.9\linewidth]{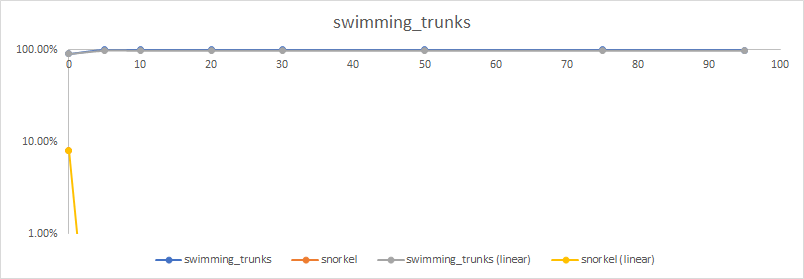}
        \includegraphics[width=.9\linewidth]{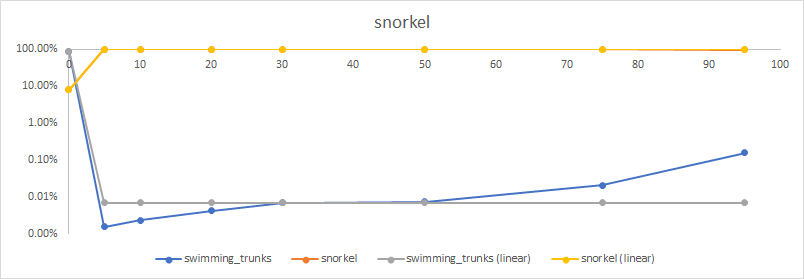}
       \caption{90.10\% \texttt{swimming\_trunks}, 8.03\% \texttt{snorkel}}
    \end{subfigure}
     \begin{subfigure}[t]{.5\textwidth}
        \centering
        \includegraphics[width=.9\linewidth]{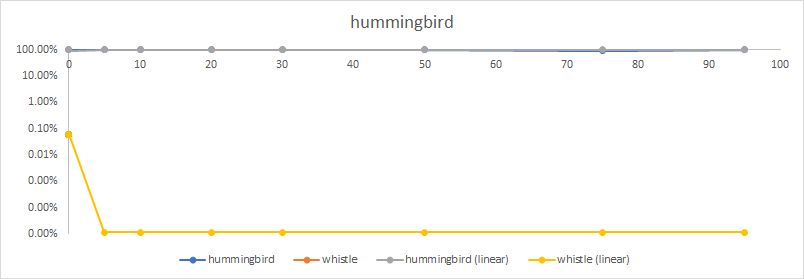}
        \includegraphics[width=.9\linewidth]{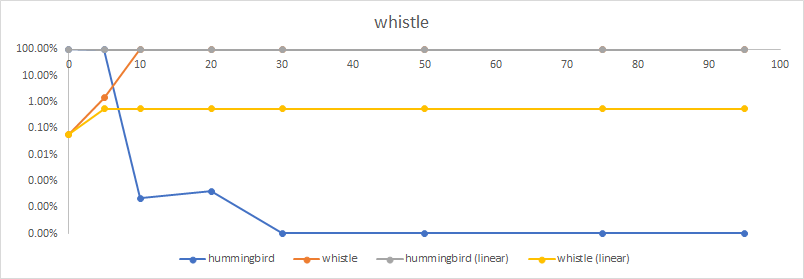}
       \caption{99.85\% \texttt{hummingbird}, 0.06\% \texttt{whistle}}
    \end{subfigure}
     \begin{subfigure}[t]{.5\textwidth}
        \centering
        \includegraphics[width=.9\linewidth]{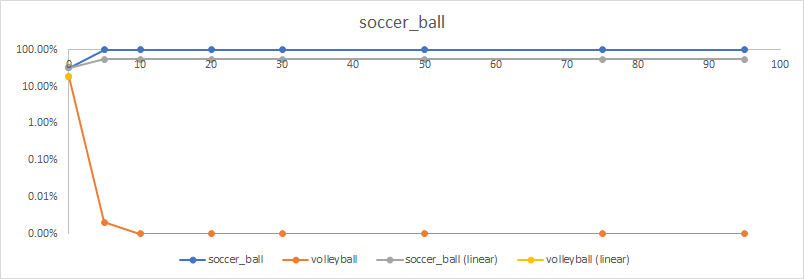}
        \includegraphics[width=.9\linewidth]{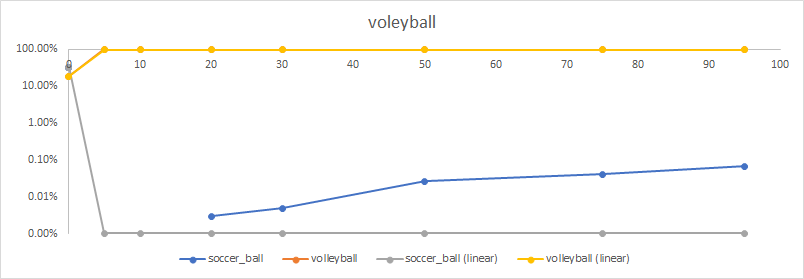}
       \caption{32.85\% \texttt{soccer\_ball}, 18.45\% \texttt{volleyball}}
    \end{subfigure}
     \begin{subfigure}[t]{.5\textwidth}
        \centering
        \includegraphics[width=.9\linewidth]{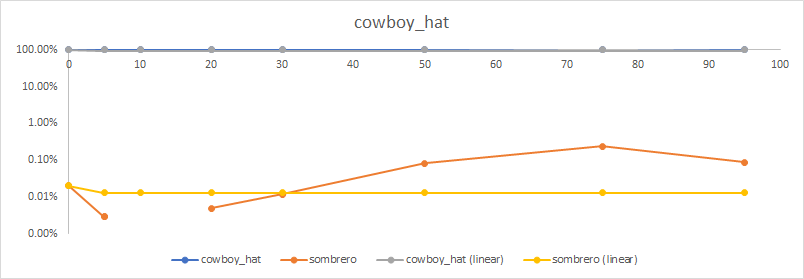}
        \includegraphics[width=.9\linewidth]{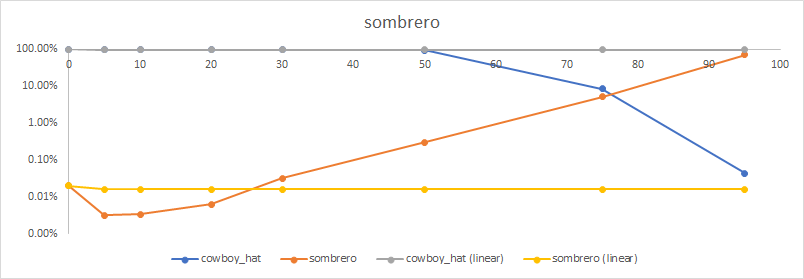}
       \caption{99.99\% \texttt{cowboy\_hat}, 0.01\% \texttt{sombrero}}
    \end{subfigure}
     \begin{subfigure}[t]{.5\textwidth}
        \centering
        \includegraphics[width=.9\linewidth]{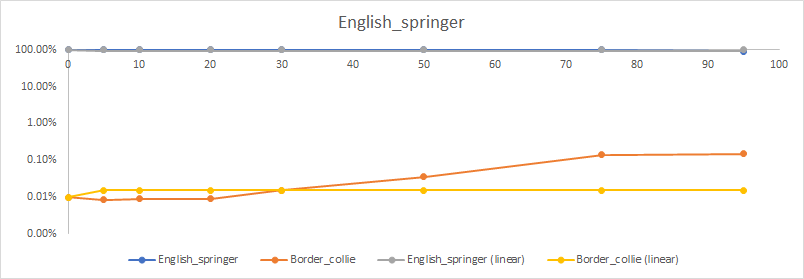}
        \includegraphics[width=.9\linewidth]{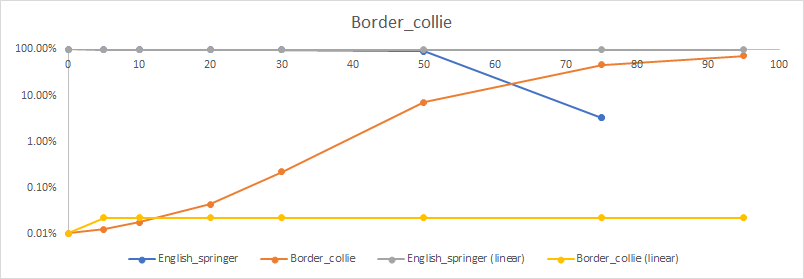}
       \caption{99.98\% \texttt{English\_springer}, 0.01\% \texttt{Border\_collie}}
    \end{subfigure}    
     \begin{subfigure}[t]{.5\textwidth}
        \centering
        \includegraphics[width=.9\linewidth]{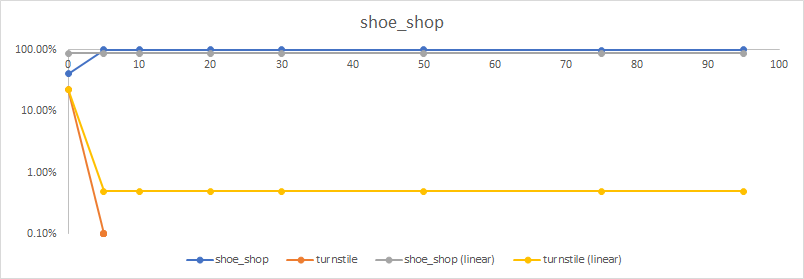}
        \includegraphics[width=.9\linewidth]{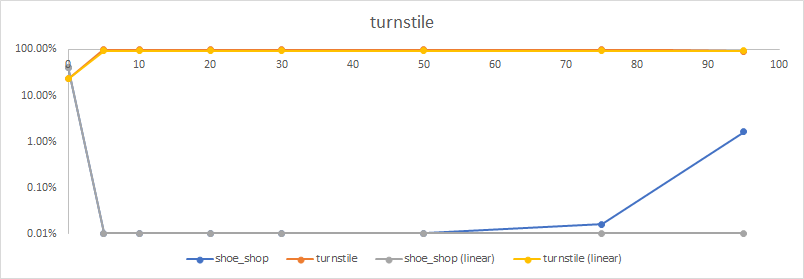}
       \caption{40.89\% \texttt{shoe\_shop}, 22.80\% \texttt{turnstile}}
    \end{subfigure}
     \begin{subfigure}[t]{.5\textwidth}
        \centering
        \includegraphics[width=.9\linewidth]{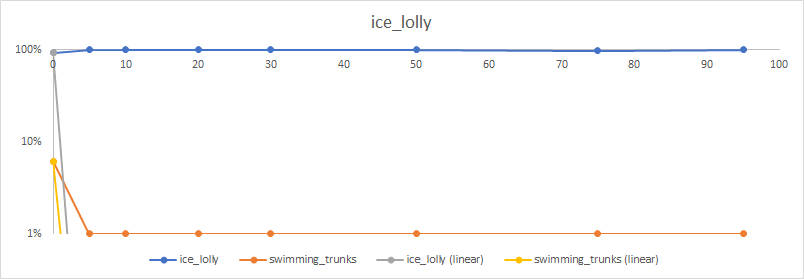}
        \includegraphics[width=.9\linewidth]{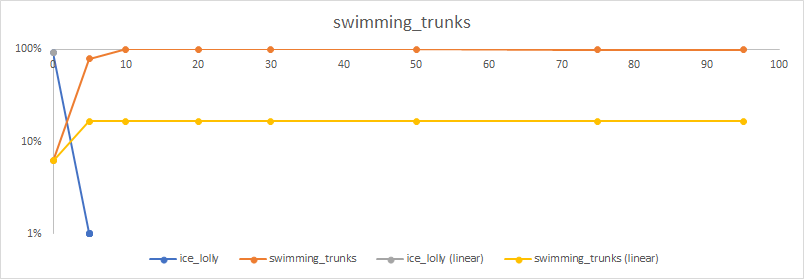}
       \caption{92.29\% \texttt{ice\_lolly},  6.14\% \texttt{swimming\_trunks}}
    \end{subfigure}
     \begin{subfigure}[t]{.5\textwidth}
        \centering
        \includegraphics[width=.9\linewidth]{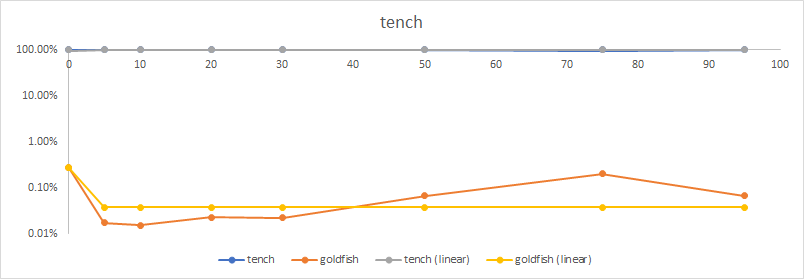}
        \includegraphics[width=.9\linewidth]{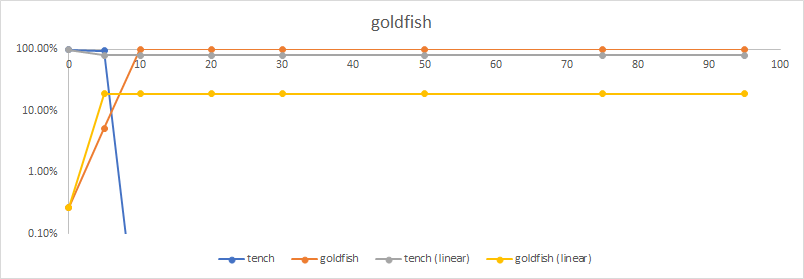}
       \caption{99.70\% \texttt{tench}, 0.27\% \texttt{goldfish}}
    \end{subfigure}
\caption{Optimization trajectories for image modifications from Figure \ref{fig:images}. Graph titles are the names of classes being optimized, x-axes are \% of pixels blurred, and y-axes are resulting class probabilities.}\label{fig:graphs}
\end{figure}

Looking at optimization trajectories in Figure \ref{fig:graphs}, it is apparent that not all trajectories are equal. Some trajectories reach the pure class representation almost immediately from the original image, indicating that the initial gradient, calculated at the original image, is sufficient, and that nonlinearities of the DNN are relatively unimportant at that point. Other trajectories, on the other hand, only find a path to the pure class representation once 50\% or 75\% of the image has been blurred out. For these images, the nonlinearities of the DNN play a more important part in their classification, and the linear approximation fails to provide a correct interpretation. 

More specifically, is apparent from the optimization trajectories that there is a threshold in the proportion of the image that has to be blurred before the maximum is reached. The blurring operator serves to destroy competing classes, thus enabling the target class to dominate the classification. The threshold corresponds to the proportion of the image  contributing to the competing classification. Only when the competing classifications have been sufficiently annihilated by the blurring operator, the target class is able to come through. 

The threshold value is not directly related to the class probability in the original classification. This is apparent in the examples of, on one hand, (h) \texttt{goldfish} ($p=0.27\%; \texttt{thr} = 5-10\%$) and (b) \texttt{whistle} ($p=0.06\%; \texttt{thr} = 5-10\%$) , and on the other (e) \texttt{Border\_collie} ($p=0.01\%; \texttt{thr} = 50-75\%$) and (d) \texttt{sombrero} ($p=0.01\%; \texttt{thr} = 75-95\%$) -- different images with class probabilities of the order of $0.1\%$ or less, but with thresholds ranging from $0$ to $95\%$. Instead, as expected, the threshold value appears to be directly linked to the convexity of the DNN at the relevant point.

\begin{figure}[ht]
        \centering
        \includegraphics[width=0.2\linewidth]{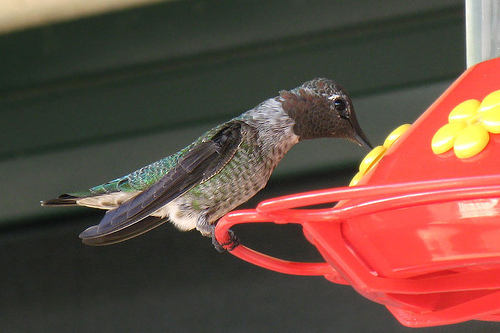}
        \includegraphics[width=0.2\linewidth]{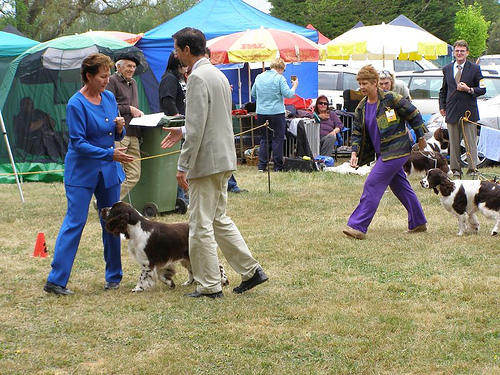}     \\    
        \texttt{hummingbird} (94) \quad \texttt{whistle} (902) \quad \texttt{English\_springer} (217) \quad \texttt{Border\_collie} (232) \\
        \includegraphics[width=0.2\linewidth]{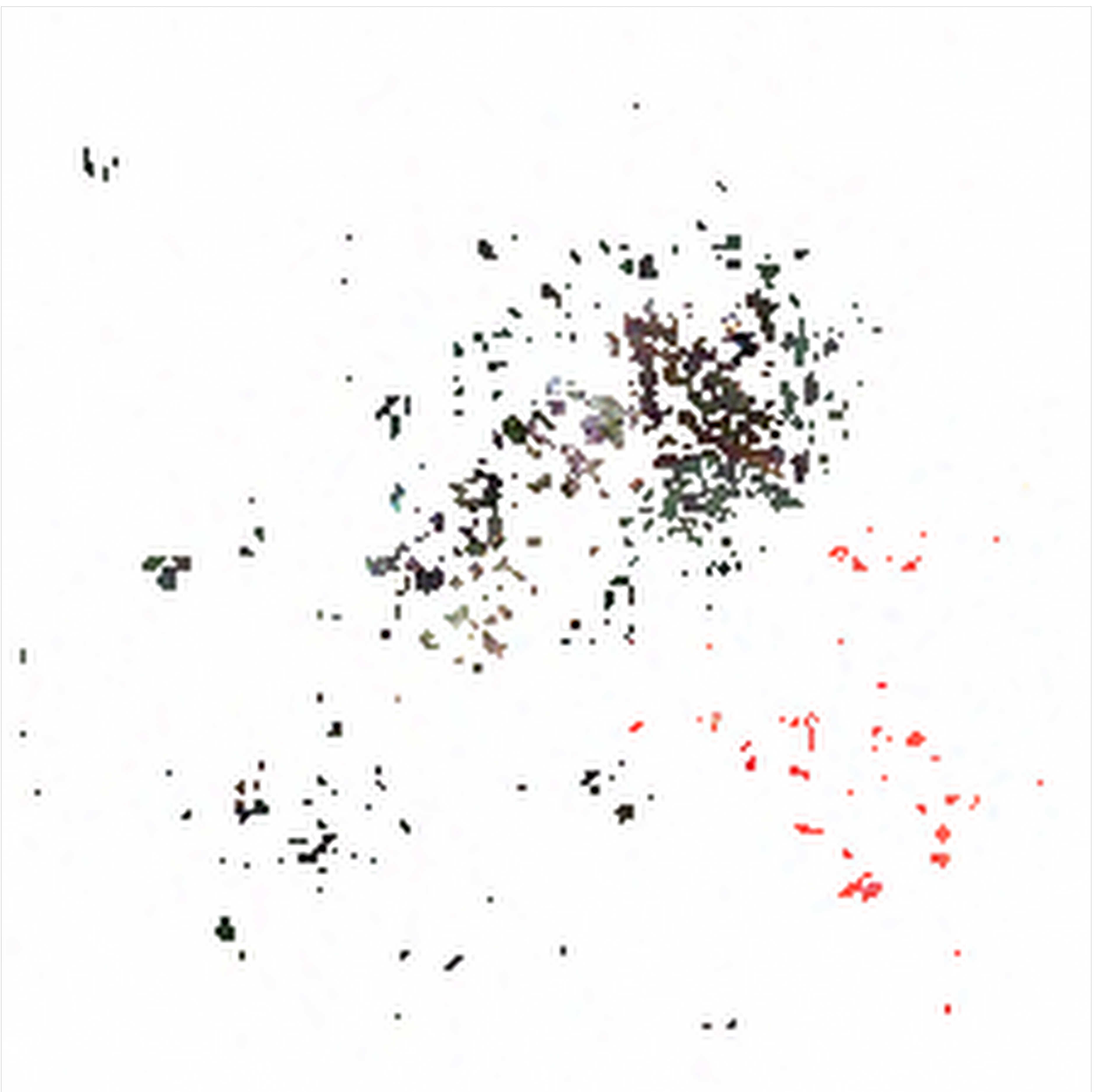}
        \includegraphics[width=0.2\linewidth]{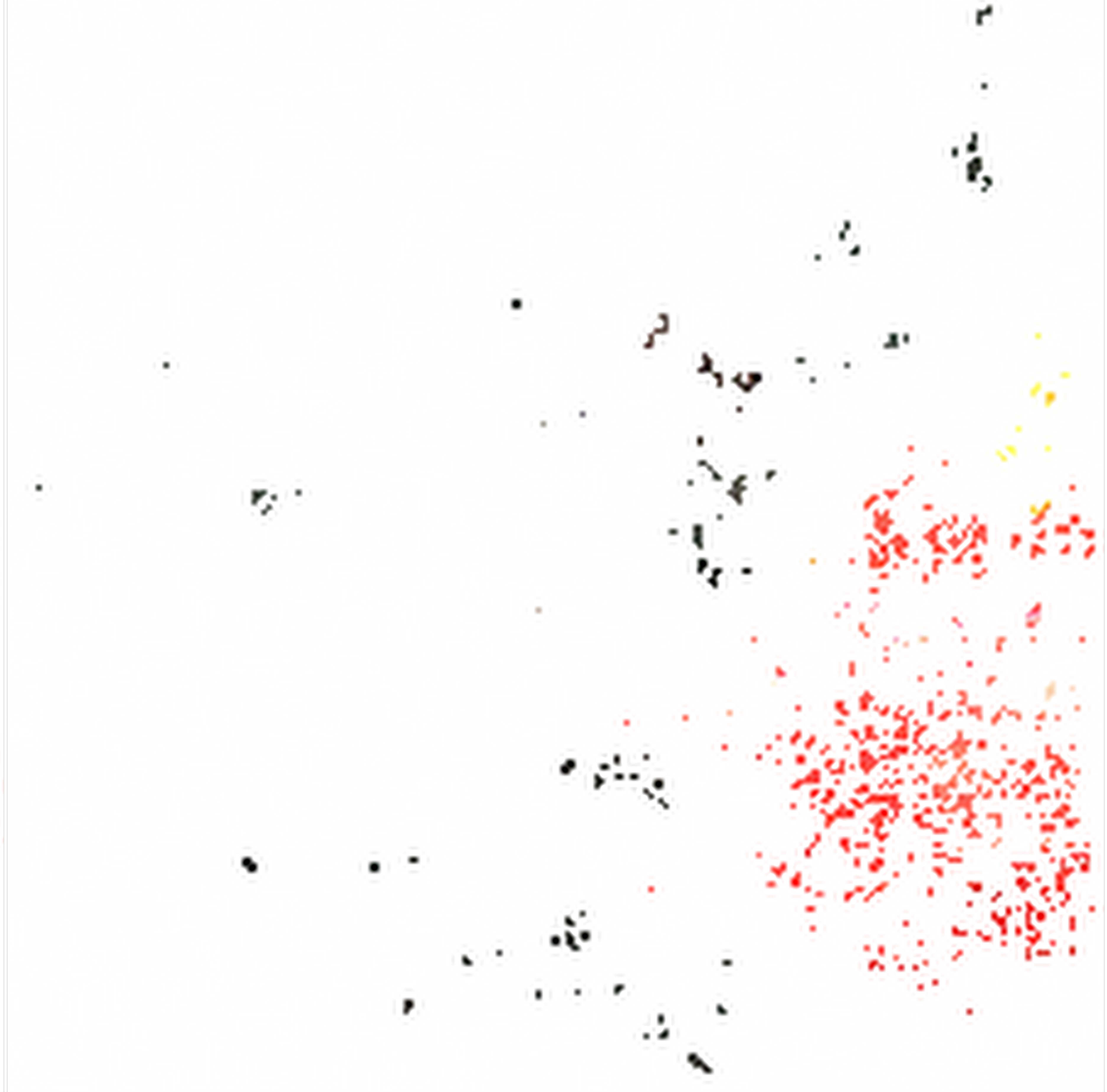}        \includegraphics[width=0.2\linewidth]{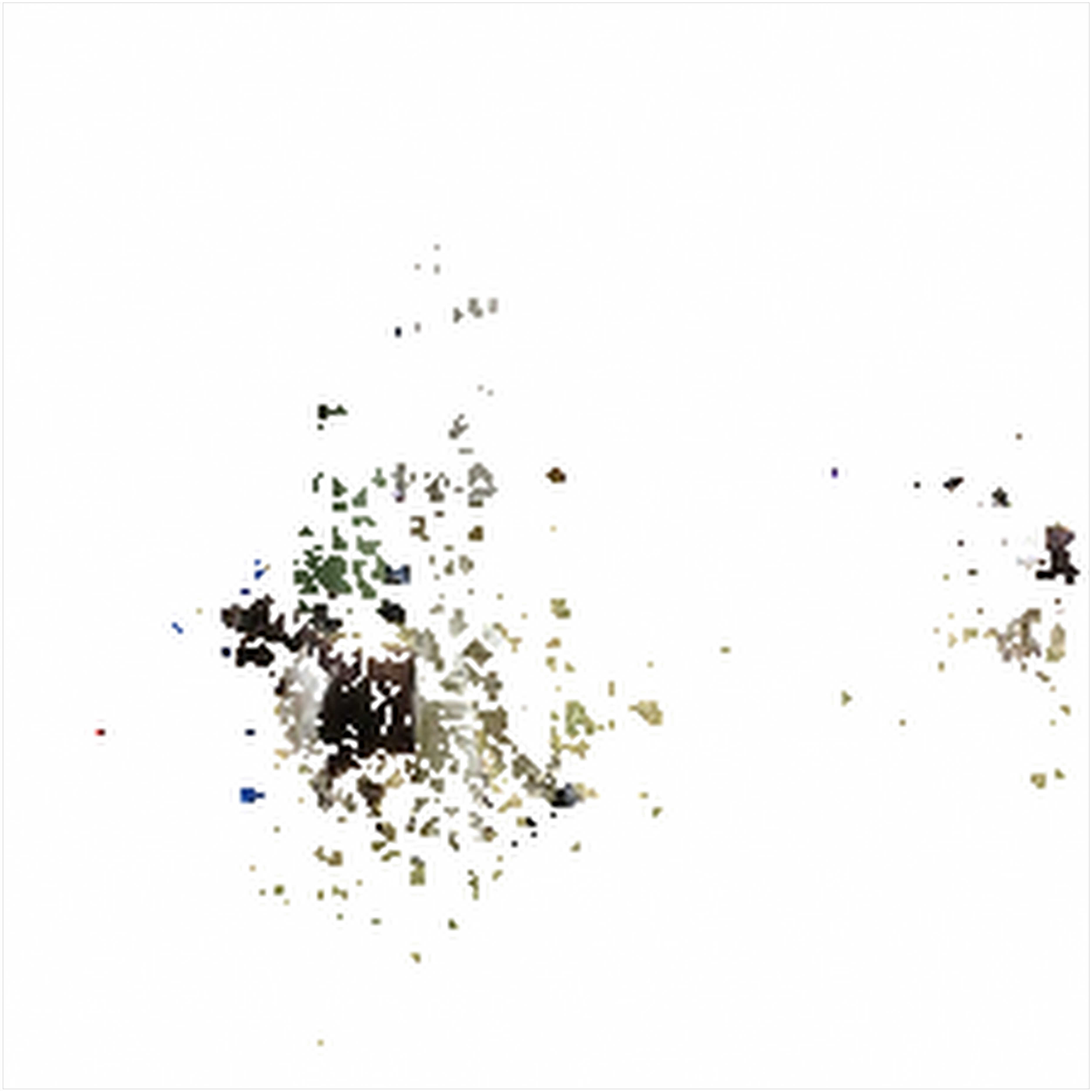}
        \includegraphics[width=0.2\linewidth]{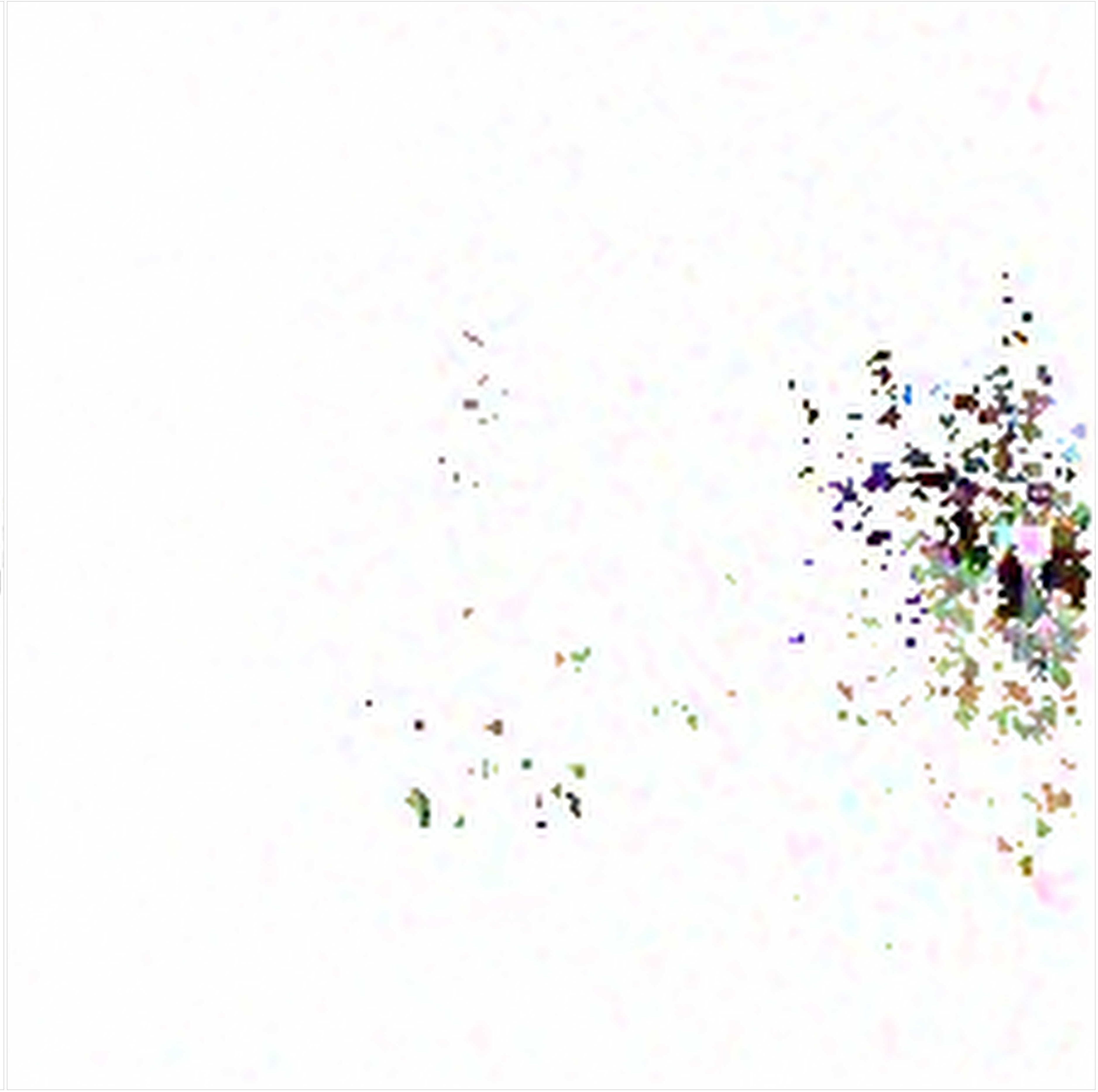}
        \includegraphics[width=0.2\linewidth]{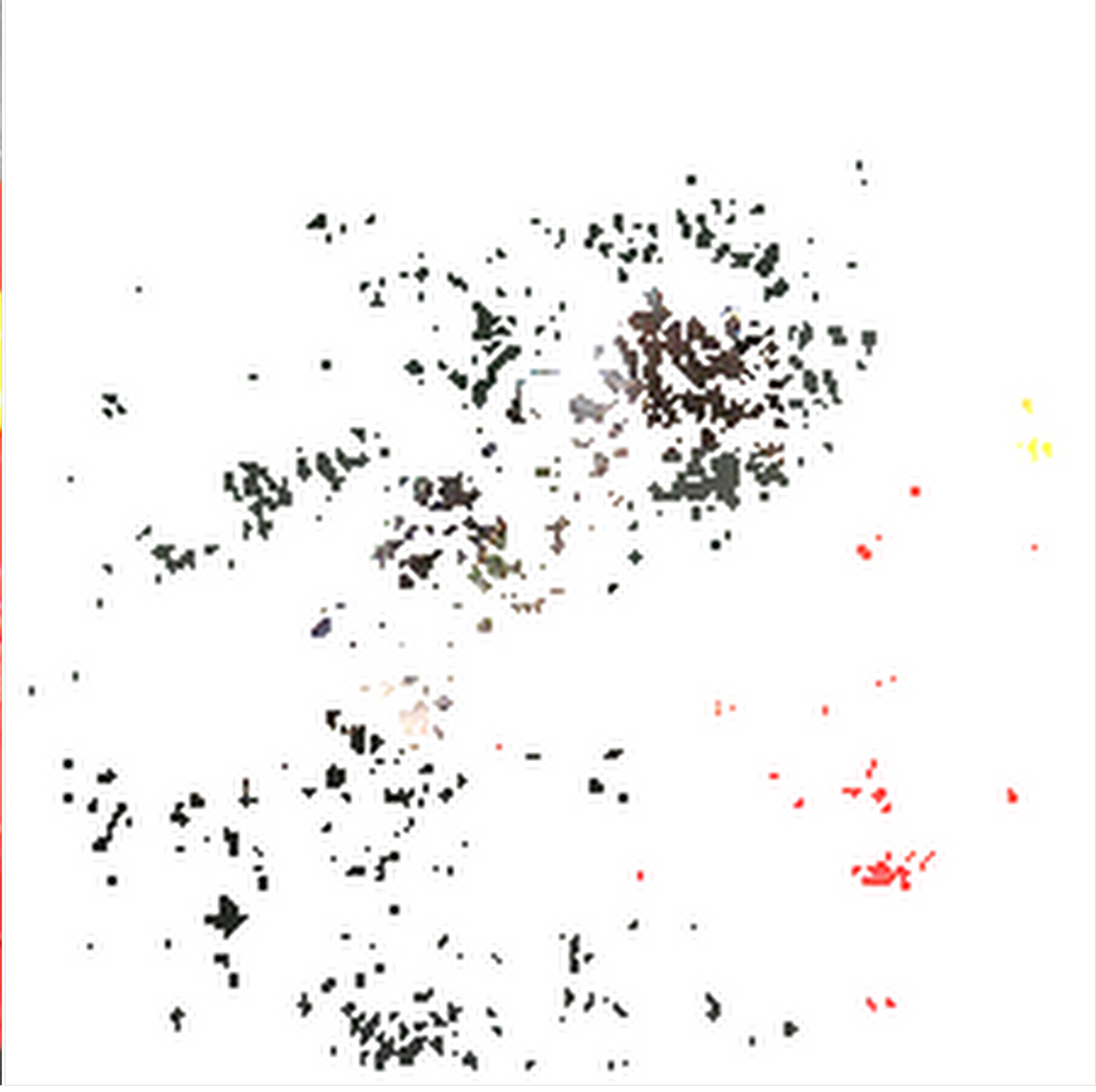}
        \includegraphics[width=0.2\linewidth]{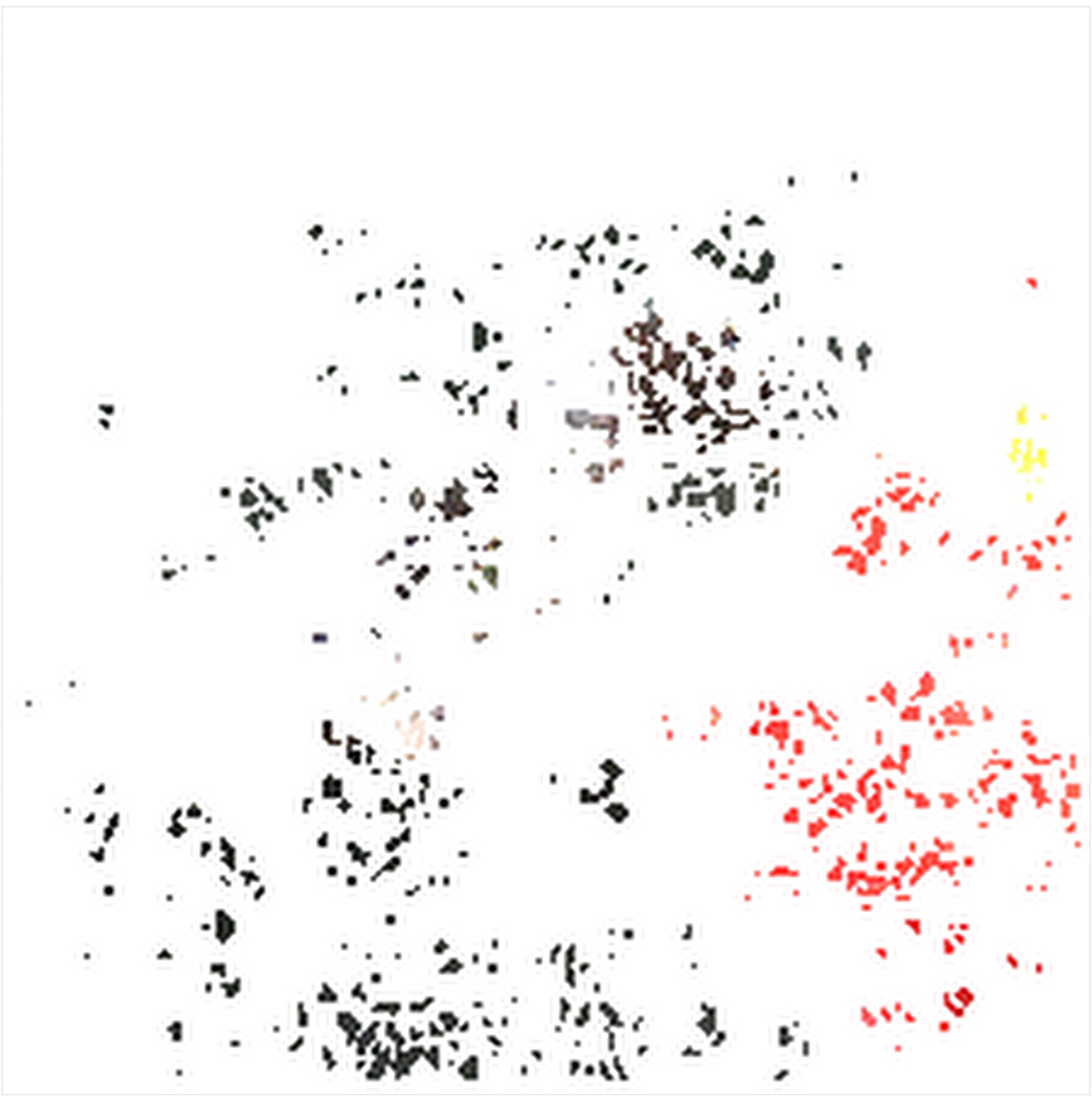}        \includegraphics[width=0.2\linewidth]{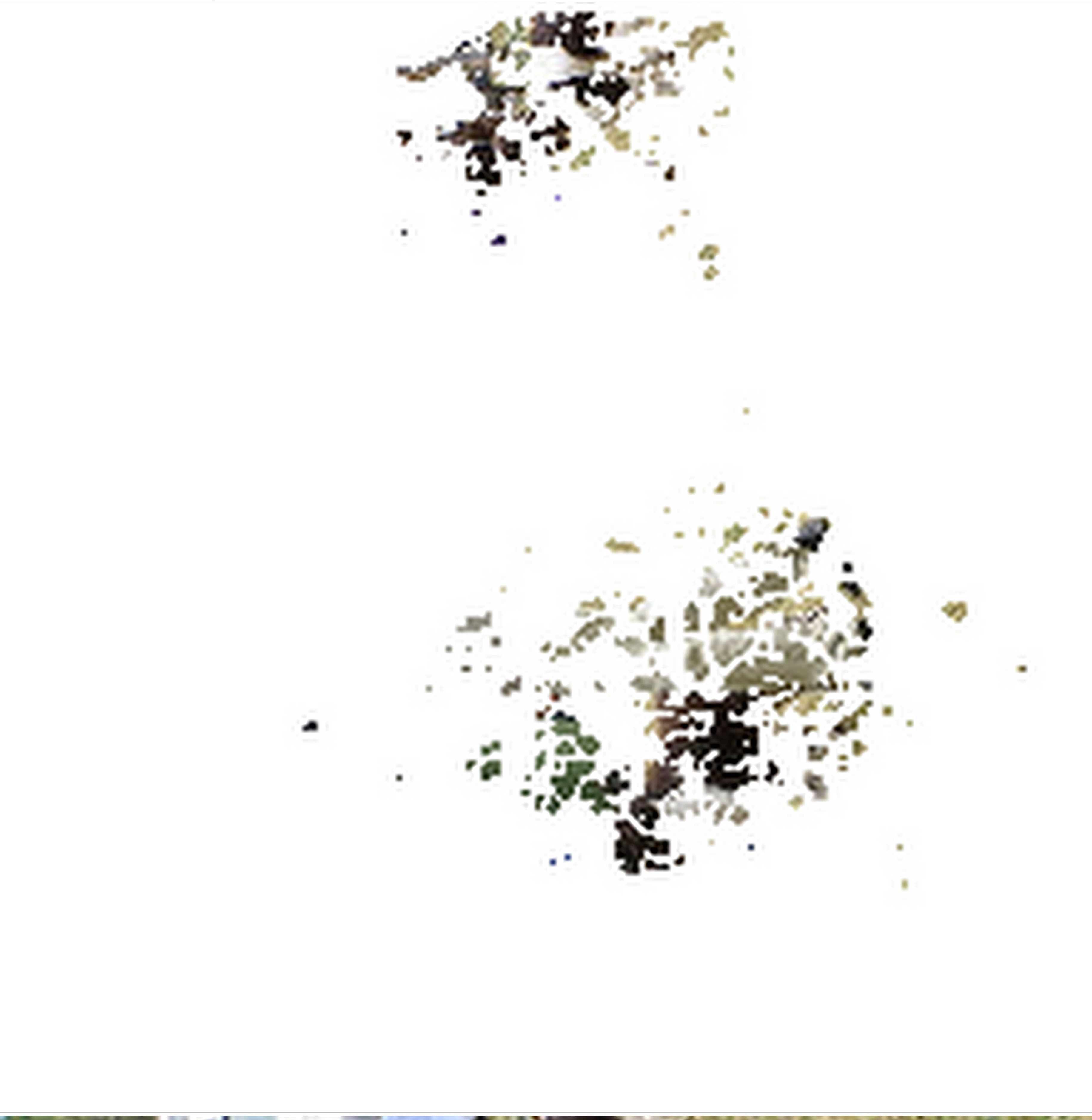}
        \includegraphics[width=0.2\linewidth]{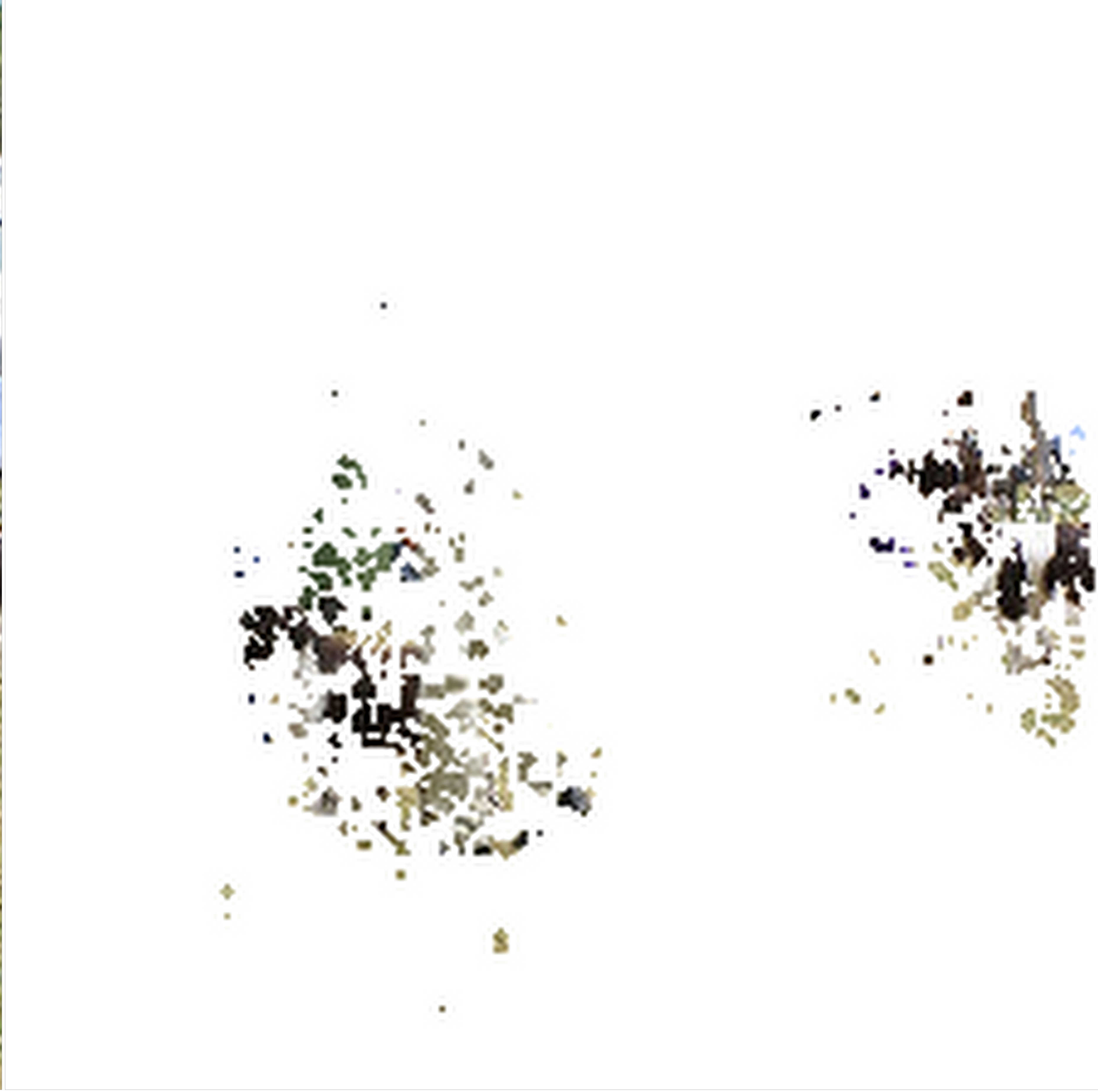}
        \includegraphics[width=0.2\linewidth]{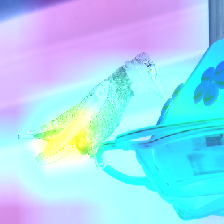}
        \includegraphics[width=0.2\linewidth]{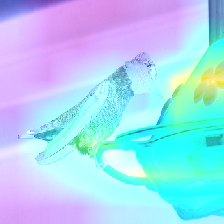}        \includegraphics[width=0.2\linewidth]{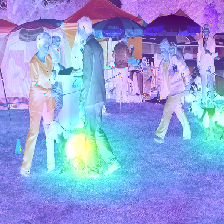}
        \includegraphics[width=0.2\linewidth]{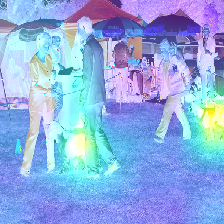}  
        \includegraphics[width=0.2\linewidth]{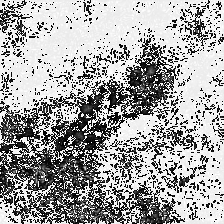}
        \includegraphics[width=0.2\linewidth]{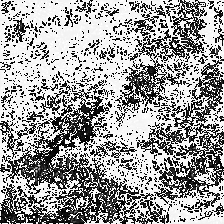}        \includegraphics[width=0.2\linewidth]{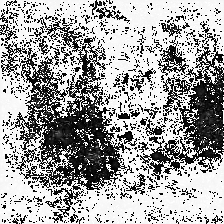}
        \includegraphics[width=0.2\linewidth]{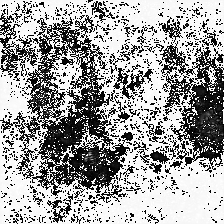}         
        \includegraphics[width=0.2\linewidth]{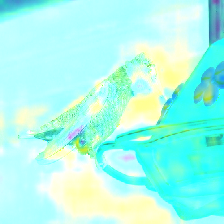}
        \includegraphics[width=0.2\linewidth]{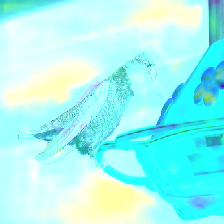}  \includegraphics[width=0.2\linewidth]{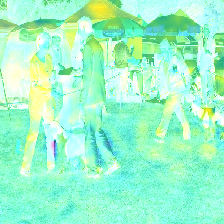}
        \includegraphics[width=0.2\linewidth]{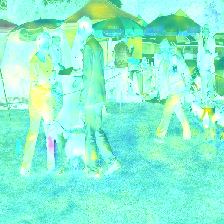} 
 \caption{Images from Figure \ref{fig:images}(b) and (e) with different saliency maps for their respective top two classes.\\
 The explainers are, from top: Nonlinear (95\% masked), (linear) Vanilla Gradient, GradCAM, SmoothGrad and Occlusion Senistivity (4 patches).} \label{fig:maps}
\end{figure}

To verify this, we looked at direct comparisons between linear and nonlinear saliency maps in more detail. The comparison between the linear and nonlinear vanilla gradient saliency maps for two tentatively nonlinear examples (b) and (e) from Figures \ref{fig:images} and \ref{fig:graphs}, is shown in Figure \ref{fig:maps}. 

The image from example (e) is a fairly complex image containing two people and a dog in the front left of the image, another person and a dog to the back right, and several more people, parasols, cars and marquees in the background. The Xception DNN identified the image as being 99.98\% \texttt{English\_springer}, 0.01\% \texttt{Border\_collie}.

All the saliency maps correctly identified the two dogs in the image as being crucial to the classification. Linear saliency maps for either the \texttt{English\_springer} or \texttt{Border\_collie} classes select both dogs more or less equally at the masking ratio of 95\%. According to the linear saliency maps, both dogs are roughly equally responsible for both classes, and the DNN is pretty clear that they are both English Springer Spaniels, or, significantly less likely, Border Collies. 

Nonlinear saliency maps, however, show something quite different. While largely agreeing with linear saliency maps at low masking ratios, at 95\% masking ratio they clearly show that the dog in the front left of the picture is identified as an English Springer Spaniel, while the dog in the back right is identified as a Border Collie. As it happens, the DNN is probably wrong on the latter point -- to the untrained eye of the authors, the dog in the back right also appears to be an English Springer Spaniel. Whether the DNN is right or wrong is, however, in this particular context immaterial. The relevant point is that this classification can't be correctly interpreted without accounting for the local nonlinearity of the DNN through nonlinear saliency maps.

A similar pattern is seen in example (b). The image shows a hummingbird sat on a bird feeder. In this case, both linear and nonlinear saliency maps correctly identify the hummingbird, blurring out the bird feeder. On the second category, \texttt{whistle}, they, however, differ. The linear saliency map selects both the bird and the feeder as contributing to \texttt{whistle}, while the nonlinear saliency map selects only the feeder, masking the bird.

We can therefore conclude that nonlinear saliency maps do generally provide a different interpretation from that of linear saliency maps. In our randomly chosen examples, the interpretation given by nonlinear saliency maps appears to be more sensible than that given by linear saliency maps.

For comparison, we have included four further state-of-the art saliency map techniques for the same images and the same classes; namely, GradCAM \cite{gradcam}, SmoothGrad \cite{smoothgrad} and Occlusion Sensitivity \cite{occlusion}, all as implemented in the python package \texttt{tf-explain} \cite{tfexplain}.

Out the tested methods, GradCAM was the only one to identify separate parts of the images as pertaining to the hummingbird versus the whistle, and the English springer vs the border collie. Its specificity on the border collie was comparable to that of vanilla linear gradient, identifying the dog on the left as the English springer, and both dogs as border collies. On the hummingbird / whistle, GradCAM outperformed the vanilla linear gradient, and it was comparable to the nonlinear gradient. Neither SmoothGrad nor Occlusion Sensitivity showed a meaningful interpretation of the second-ranked class.  

\begin{figure}[ht]
        \centering
        \includegraphics[width=1\linewidth]{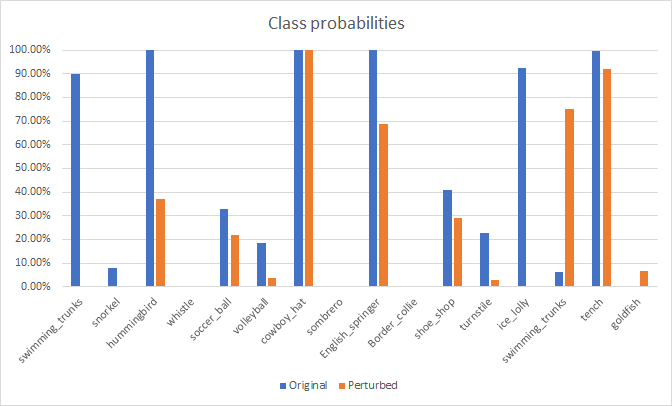}
\caption{Effects of selectively blurring nonlinearly salient pixels on the relevant class probabilities.} \label{fig:pert}
\end{figure}

To further elaborate on this point, we have selectively blurred the $95\%$-salient pixels for each relevant class probabilities; that is, for example, in Figure \ref{fig:images}, we have blurred the selected pixels that correspond to $95\%$ saliency for \texttt{swimming\_trunks}, and looked at how this affects the class probability for the class \texttt{swimming\_trunks}. This was repeated for top two classes for each of the images. The results are shown in Figure \ref{fig:pert}.

As can be seen from Figure \ref{fig:pert}, selectively blurring nonlinearly salient pixels results in significantly lowering any class probability; the only exception to that are the classes \texttt{swimming\_trunks} and \texttt{goldfish} in Figure \ref{fig:images} g) and h), which are increased by such blurring. It is notable that neither of these two classifications rely on fine-scale detail, which might explain why blurring has the opposite of the desired effect.

Finally, we note that the optimization algorithm we used is a simple extension of gradient descent, which is a relatively unsophisticated optimizer. Different core optimization methods  generally generate somewhat different trajectories. The direction of image enhancement is, however, only one part of the optimization procedure, with the other being the filtering operation. As noted above, the critical value governing the speed of convergence is the threshold in the proportion of image that is blurred, which is only weakly linked to the direction of image enhancement. Changing the optimizer therefore generally does not significantly affect the critical filtering threshold after which the maximum is reached, and it does not have a strong effect on the speed of convergence.

How important are nonlinear saliency maps for interpretability of machine learning models? That is, of course, at this stage still unclear. It is clear that fully accounting for nonlinearities in the model is of considerable importance for the task of interpretability. The class of methods that we describe is a step forward in this direction. It is able to do the same job as linear saliency maps where linear saliency maps are correct, and it provides an important, distinct change of direction where they are not.

The examples we presented come from computer vision and image classification, which is admittedly a task that often crosses the boundary into being to some extent subjective. Assigning complex images to simple, single classes is always somewhat arbitrary. The example image of two dogs could equally validly be classified as depicting people, a fair, or a lawn. 

The true test of the methods presented here will come in real life applications which fall outside the scope of this study, such as medical image classification, material defect analysis and others.

\section*{Methods}

All calculations were conducted using Tensorflow 1.18 (\texttt{http://www.tensorflow.org}), \texttt{tf-explain} \cite{tfexplain} and NumPy (\texttt{https://numpy.org}) running on Python 3.6 (\texttt{http://python.org}). 

We used pretrained tensorflow models (\texttt{https://tfhub.dev/}) corresponding to the well documented DNNs such as VGG16, ResNet15, MobileNet, InceptionV3 and Xception.

Each image was classified by the DNN models. 

Instead of masking non-highlighted pixels, we subject them to a low-pass filter, effectively blurring the non-salient parts of the image. 

We then enhance the highlighted pixels in the direction of the gradient. By iteratively repeating this step, we are able to selectively amplify the features that are important to classifying the image into any chosen class. The corresponding saliency map is then obtained by masking any blurred pixels.

Neither of those steps is sufficient in its own right; blurring pixels without gradient enhancement only increases the entropy of the image, effectively blurring the image to the extent that it can not be reliably classified as anything. Gradient enhancement without blurring, on the other hand, reaches the boundaries of the RGB box and stops there, without being able to add significantly to the classification.

The nonlinear saliency maps corresponding to any class was calculated by:
\begin{enumerate}
\item selecting the appropriate column for the desired class from the Jacobian of the output vecor with respect to the input tensor
\item for a given threshold, updating the image according to the logic
\begin{verbatim}
next_img =  np.clip(
       blur_below_level(prev_img, sensitivity_norm, L) + 
       littleH * sensitivity, 0., 1.)
\end{verbatim}
\item obtaining the saliency map using the related logic
\begin{verbatim}
next_map =  np.clip(
       mask_below_level(prev_img, sensitivity_norm, L) + 
       littleH * sensitivity, 0., 1.)
\end{verbatim}
\end{enumerate}
Here, \texttt{sensitivity} refers to the relevant column of the Jacobian,  \texttt{sensitivity\_norm} to its pixel-wise norm (Euclidean norm over the RGB channel direction),  \texttt{blur\_below\_level} performs the low pass filtering on pixels where \texttt{sensitivity\_norm} falls below  the selected level \texttt{L}, \texttt{mask\_below\_level} covers the same pixels in a background colour (in all our examples, white) and \texttt{littleH} is a small number controling the size of the optimization step. The clipping operation ensures that all RGB values remain within the RGB range. 

The level parameter threshold \texttt{L}, and the distance parameter \texttt{littleH} control the step size in the optimization process.

\section*{Data Availability}

The datasets analysed during the current study are available in the Imagenet ILSRVC 2017 repository, \texttt{https://image-net.org}.

\section*{Author Consent}

The authors consent for publication of identifying information/images in an online open-access publication.

\pagenumbering{gobble} 

\end{document}